# EXACT REASONING ABOUT UNCERTAINTY

Samuel Holtzman
Jack Breese
Department of Engineering-Economic Systems
Stanford University
Stanford, California 94305

June, 1985

## ABSTRACT

This paper discusses the use of uncertain knowledge in expert systems. The discussion is focused on the use of these systems to support decision making in complex, uncertain environments. Within this context, our research indicates that strictly probabilistic representations, which enable the use of decision-theoretic reasoning, are highly preferable to recently proposed alternatives. Furthermore, the language of influence diagrams and a corresponding methodology that allows decision theory to be used effectively and efficiently as a decision-making aid is discussed. A system, called RACHEL, that helps infertile couples select pertinent medical treatments provides an illustration of this methodology.

## I. INTRODUCTION: MAKING DIFFICULT DECISIONS

Reasoning about uncertain knowledge is fundamentally a question of developing and manipulating a measure of our information about various possible worlds. We consider reasoning about uncertainty in the context of decision making, i.e. when there is an explicit intention to irrevocably allocate valuable resources. This contrasts with a situation where one is "curious" about the world, or wishes to extract patterns or frequencies from a large body of information with no direct intention to make a decision.

Consider a situation where the decision maker is unable to select a clear course of action. This inability is most often indicative of the absence of a satisfactory representation (model) of the decision, or of an acceptable method for reasoning with this representation to yield a recommendation for action. In particular, the following situations are often associated with difficult decision making.

- Dissatisfaction with Available Alternatives
    The decision maker feels that there must be other choices
    beyond those being considered.

- Lack of Clarity
    One or more elements of the decision model are ambiguously
    defined.

- Inadequate Structural Understanding
    The relationships and dependencies among the variables in the
    decision model are not clearly represented.

- Inferential Complexity
    The decision context requires a major inferential effort to deduce
    the effect of actions on outcomes.

- Combinatorial Complexity
    The number of possibilities being considered is overwhelmingly large.

- Inability to Deal with Uncertainty
    Lack of a satisfactory means for representing, measuring, and
    reasoning with information about the possible outcomes of
    decision factors.

In light of these common obstacles in the decision-making process, an individual facing a difficult decision often needs assistance in four areas.

- Generating Alternatives
    That is, developing possible courses of action.

- Constructing a Well-defined Decision Model
    That is, identify the factors which impinge on the decision and define
    the relationships between these factors.

- Focusing Attention on the Major Issues of the Decision
    That is, distinguishing what is important to the individual
    decision maker from what is generally relevant to all problems
    within the decision context.

- Combining the Available Information into a Decision Recommendation
    That is, reasoning with the decision model to obtain a recommended
    course of action.

All but the first of these four needs for decision support is discussed.

## II. EXPERT SYSTEMS AS DECISION AIDS

Implicit in the use of computer-based techniques for decision making, and of expert decision support systems in particular, is the contention that the effectiveness of decision making can be improved through the use of formal analytic techniques. Specifically, systems that provide a recommendation for action must have within them one or more formal statements that conclude a suggestion for action on the basis of a formal representation of the decision. We refer to such statements, which can be implicit as well as explicit, as "action axioms" [Holtzman, 1984].

An essential requirement for successful decision support systems is that they be based on an action axiom which is explicit and can be clearly presented to the decision maker for approval. Ultimately, the value of the system's recommendations hinge on the extent to which the decision maker trusts the system's reasoning paradigm. The choice of an underlying normative axiomatic system can have a profound effect on the usefulness of an expert decision support system.

An evaluation of the available representations for uncertain knowledge from a decision perspective reveals the desirability of strict probabilistic forms. Beyond the classic works of Savage [1972] and de Finetti [1968], who argue eloquently for the formalism of decision theory as a basis for normative reasoning, Lindley [1982] has shown that, under very weak assumptions, non-probabilistic representations of uncertainty (including fuzzy sets [Zadeh, 1981] and Dempster-Shafer evidence measures [Shafer, 1976],) are theoretically inadmissible. In this context, inadmissibility refers to the existence of legitimate situations where the decision maker would be given a recommendation to act in a way that would decrease his or her welfare for certain.

In addition to its admissibility, there is a more pragmatic argument for basing expert decision support systems on a decision-theoretic action axiom, that is, on the maximization of expected utility. Extensive experience in the professional practice of decision analysis provides a mature and well-tested methodology for the formulation, assessment, and use of decision-theoretic models. Effective techniques exist for the development of individual decision models and for the assessment of the values of model parameters. These techniques are well understood and can be almost directly incorporated into computer-based decision tools. To our knowledge, no counterpart to this methodology exists for non-probabilistic representations of uncertainty.

Designing an expert decision support system on the basis of a decision-theoretic representation directly addresses the need of decision makers to construct a well-defined decision model. Furthermore, decision theory provides a powerful means for combining the available information into a decision recommendation.

Of course, although it surely helps, a good wrench does not necessarily make a good plumber. Similarly, a good representation language paves the way to good representations, but does not guarantee them. Beyond its language, much of the value of the methodology of decision analysis lies in the way in which it facilitates the formalization of a real decision into a model.

## III. DEVELOPING REPRESENTATIONS OF SPECIFIC DECISIONS

A fundamental feature of decision analysis as a normative methodology is a recognition of the fact that the most common major stumbling block in making a decision lies not in solving an existing decision model but in formulating such a model. A useful way to visualize the decision analytic process from this perspective is as an interactive three-stage process illustrated on the following page. Foreshadowing some of our later discussion, we note that the stages of this process have a direct correspondence with three major functions of an expert system: theory formation, inference, and explanation. The three stages of the decision analysis process can be described as follows.

1. Formalizing the decision, that is, building a formal model of a real decision situation -- theory formation --,

2. Solving the decision model, that is, applying a set of formal axioms to the decision model to deduce its implicit recommendation -- inference --, and

3. Clarifying the decision recommendation, that is, interpreting the model solution in terms of its implications as a guide for real action -- explanation.

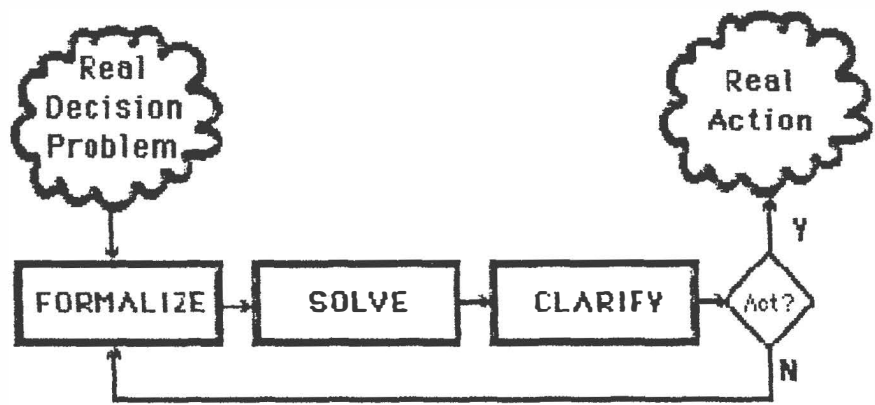

A three-stage decision process

To be useful, a decision methodology must help the decision maker identify the most significant features of his decision situation. It is much more common for a decision to be difficult because too much is being considered than because too few issues have been raised. One could argue that this prevalence of overly complex models is evidence that many of our decision-making habits are a legacy of a time when information, rather than attention, was scarce. In any case, it is simple to show that people are excellent at listing issues that are relevant to just about any decision, often in spite of their lack of direct familiarity with the decision context. In contrast, even the most experienced among us is demonstrably quite inept at identifying which of those relevant issues are important to a particular decision maker.

Much of the weight of the foregoing argument lies on the meaning we give to the terms "significant", and "important". In this discussion, these terms are essentially equivalent, and refer to the degree to which a feature in the decision model affects the decision maker's welfare and choices. As an illustration, suppose you are considering the purchase of an automobile. Assume that you are concerned about fuel efficiency and that you do not particularly care to do your own repairs. In this case, although both are relevant to your decision, the standard mileage ratings of the cars you are considering are likely to be important, whereas the fact that a car is designed using English or metric measurements is probably unimportant. A model of your decision has a better chance of being useful to you if it plainly ignores the whole issue of engine dimensions and concentrates on what you care about (such as gas mileage).

Therefore, in designing decision models one must be guided by the decision maker's preferences to be able to distinguish and focus attention on what is important, rather than on all that is relevant. A fundamental flaw in most designs for expert decision support systems is the reliance on decision models which have been built without a clear decision context. In particular, these a-priori models fail to account for the decision maker's preferences. Although valid for strictly descriptive systems and situations with very clear dominating solutions, such models have little general normative power.

Central to this building process is the use of sensitivity analysis calculations (of various sorts) as a means of focusing attention, i.e. directing search. Sensitivity analysis should be an integral part of the model-building effort -- guiding it to what is most important to the decision maker -- and not just an afterthought to check the validity of numerical parametric assessments.

## IV. EXPERT SYSTEMS FOR DECISION ANALYSIS

Decision-analytic reasoning can be incorporated into the reasoning process of an expert system in a relatively straightforward way. A useful tool for this purpose is the language of influence diagrams [Howard and Matheson, 1984]. Influence diagrams (IDs) are a generalization of decision trees and have been empirically shown to have considerable intuitive appeal for wide classes of decision makers. In addition, they are a powerful communication tool for the participants in the decision process.

Influence diagrams are acyclic, directed graphs representing the probabilistic, logical, and informational relationships between variables in a decision model. Nodes in the graph can represent the decision-maker's decision alternatives, uncertain variables in the decision domain, and the decision maker's preferences with respect to the model variables. An arc leading to a node representing a decison variable indicates the value of the preceding node is known at the time that the decision will be made. Presence of an arc between nodes representing uncertain variables indicates that the two variables may be probabilistically dependent.

A full description of IDs, their use as a decision modeling language, and their solution are beyond the scope of this paper. Relevant discussions can be found in the literature [Howard and Matheson, 1984; Olmsted, 1983; Shachter, 1984; Holtzman, 1985]. However, it is important to note that, by construction, IDs are mathematically well-defined and can be directly used in decision-theoretic calculations [Breese and Holtzman, 1984; Shachter, 1984].

Recent research indicates that the formalism of influence diagrams is an effective means for representing decision knowledge in expert systems [Holtzman, 1985]. Decision knowledge can be encoded as a collection of portions of influence diagrams. This is used to develop an overall decision model as a side effect of the inferential process of an expert system and be solved for the optimal solution in the same formalism. This form of knowledge can guide the decision-making process and, in particular, it can focus the task of formulating a decision model effectively and efficiently. An important feature of this representation of uncertain knowledge is that it does not have the tremendous requirements of data commonly associated with a-priori probabilistic models. Furthermore, it does not impose artificial independence constraints on the data it uses.

## V. RACHEL: A PILOT-LEVEL EXPERT DECISION ANALYST

An experimental expert decision system has been built as an application of the methodology discussed in this paper [Holtzman, 1985]. The system, called RACHEL, is designed to help infertile couples select a medical course of action. RACHEL concentrates on the initial formulation of a model of the patient's decision. The resulting model is then solved to yield a decision recommendation. The user is finally left in an interactive environment where he or she can analyze, modify, and resolve the model as needed.

A typical consultation with RACHEL can be divided into three phases. The first consists of the assessment of a deterministic model of patient preferences. This model is obtained by first selecting one of several available model schemas on the basis the presence or absence of major decision features and then instantiating the model schema with patient specific information. The second phase develops a probabilistic distribution for the outcome of this preference model given each and every possible decision alternative under consideration. This distribution is almost invariably obtained by constructing a fully assessed patient-specific influence diagram. The third phase of a consultation includes the solution of the initial model developed in the first two phases and concludes by placing the user in an interactive mode, where he or she can refine the decision model.

## VI. CONCLUSIONS

Decision analytic techniques are powerful tools for the design of expert systems for decision support in complex, uncertain domains. Such systems should make explicit use of the decision maker's preferences to focus attention on the important aspects of the decision, and should represent uncertain knowledge probabilistically to ensure the admissibility of their recommendations. A probabilistic representation of uncertainty provides a normative means for making inferences in support of decision making, and influence diagrams provide a natural, computable device for building and manipulating models.